\documentclass[letterpaper, 10 pt, conference]{ieeeconf}  

\IEEEoverridecommandlockouts                              

\usepackage{booktabs}
\usepackage{subcaption}
\usepackage{url}

\usepackage{graphics}
\usepackage{graphicx}
\usepackage{cite}
\usepackage{multirow}
\usepackage{color, soul}

\usepackage{amsmath} 
\usepackage{amssymb}  

\title{\LARGE \bf
PointSeg: Real-Time Semantic Segmentation \\
Based on 3D LiDAR Point Cloud}

\author{Yuan Wang$^{1}$ \ \ \ \
        Tianyue Shi$^{2}$ \ \ \ \
        Peng Yun$^{1}$ \ \ \ \
        Lei Tai$^{1}$ \ \ \ \
        Ming Liu$^{1}$
\thanks{$^{1}$The Hong Kong University of Science and Technology,
        {\tt\small \{ywangeq, pyun, ltai, eelium\}connect.ust.hk}}%
\thanks{$^{2}$Nanjing University of Aeronautics and Astronautics
        {\tt\small iristainyue@gmail.com}}%
}

\begin{document}

\maketitle
\thispagestyle{empty}
\pagestyle{empty}

\begin{abstract}
We propose \textit{PointSeg}, a real-time end-to-end semantic segmentation method for road-objects based on spherical images.
The spherical image is transformed from the 3D LiDAR point clouds with the shape $64 \times 512 \times 5$ and taken as input of the convolutional neural networks (CNNs) to predict the point-wise semantic mask.
We build the model based on the light-weight network, \textit{SqueezeNet}, with several improvements in accuracy.
It also maintains a good balance between efficiency and prediction performance.
Our model is trained on spherical images and label masks projected from the \textit{KITTI} 3D object detection dataset. 
Experiments show that \textit{PointSeg} can achieve competitive accuracy with 90 fps on a single GPU,
which makes this real-time semantic segmentation task quite compatible with the robot applications.
\end{abstract}

\section{Introduction}
\label{sec:intro}

\subsection{Motivation}
\label{sec:motivat}
3D real-time semantic segmentation plays an important role in the visual robotic perception application, such as in autonomous driving cars.
Those robotic systems collect information from the real-time perception based on different modes of sensors and understand which and where objects are on the road.
Different applications make decisions based on different perceptions, such as camera, inertial measurement units (IMUs) and LiDAR.
LiDAR scanner is one of the essential components where we can directly get the space information.
It is also less influenced by light compared with cameras and has robust features in challenging environments.
Therefore, a fast 3D semantic segmentation method will help the robot understand the world information more directly.
In addition, computation power on an autonomous driving system is quite limited to maintain those state-of-the-art sources consuming methods. Even in the workstations, the semantic segmentation is still challenging to achieve the real-time performance.
Moreover, embedded computing devices, such as Jetson TX2 and FPGA, cannot provide the same level computation ability as those regular workstations.
Because of this, a good perception method with high accuracy, low-cost memory and compatible real-time performance has become a crucial problem, which has attracted much research attention.

Previous approaches about point clouds recognition \cite{feng2014fast} \cite{himmelsbach2008lidar} mainly rely on complicated hand-crafted features,
such as surface normal or generated descriptors, and hard threshold decision rules based on a clustering algorithm.
These approaches have two problems:
(1) hand-crafted features cost much time and the results by hard threshold decision are not suitable for productions;
(2) they can not recognize the pixel level object category as the same as the semantic segmentation, which makes it difficult to apply to some autonomous driving tasks.

\begin{figure*}[!t]
  \centering
    \includegraphics[width=0.3\textwidth]{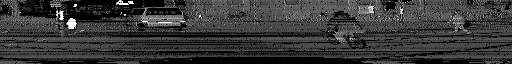}
    \includegraphics[width=0.3\textwidth]{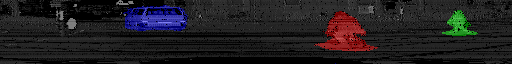}
    \includegraphics[width=0.3\textwidth]{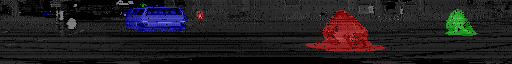}
    \begin{subfigure}{0.3\textwidth}
      \includegraphics[width=1\textwidth]{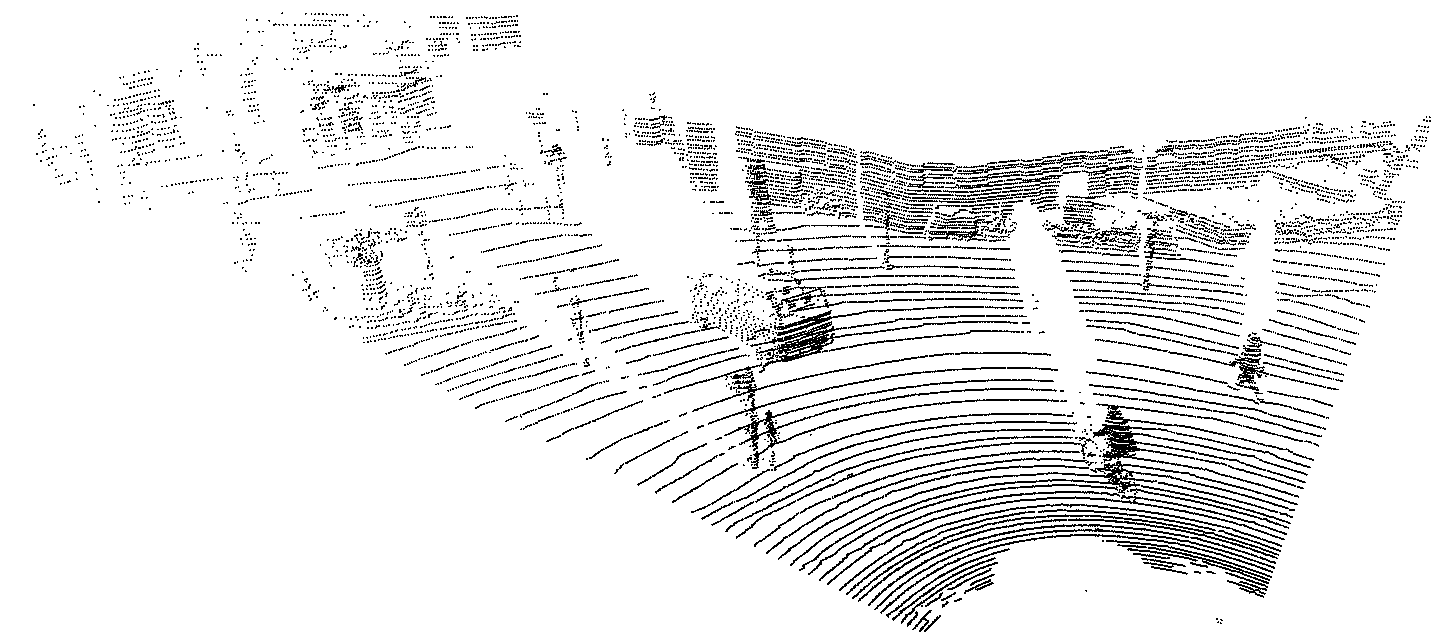}
      \caption{raw data}
    \end{subfigure}
    \begin{subfigure}{0.3\textwidth}
      \includegraphics[width=1\textwidth]{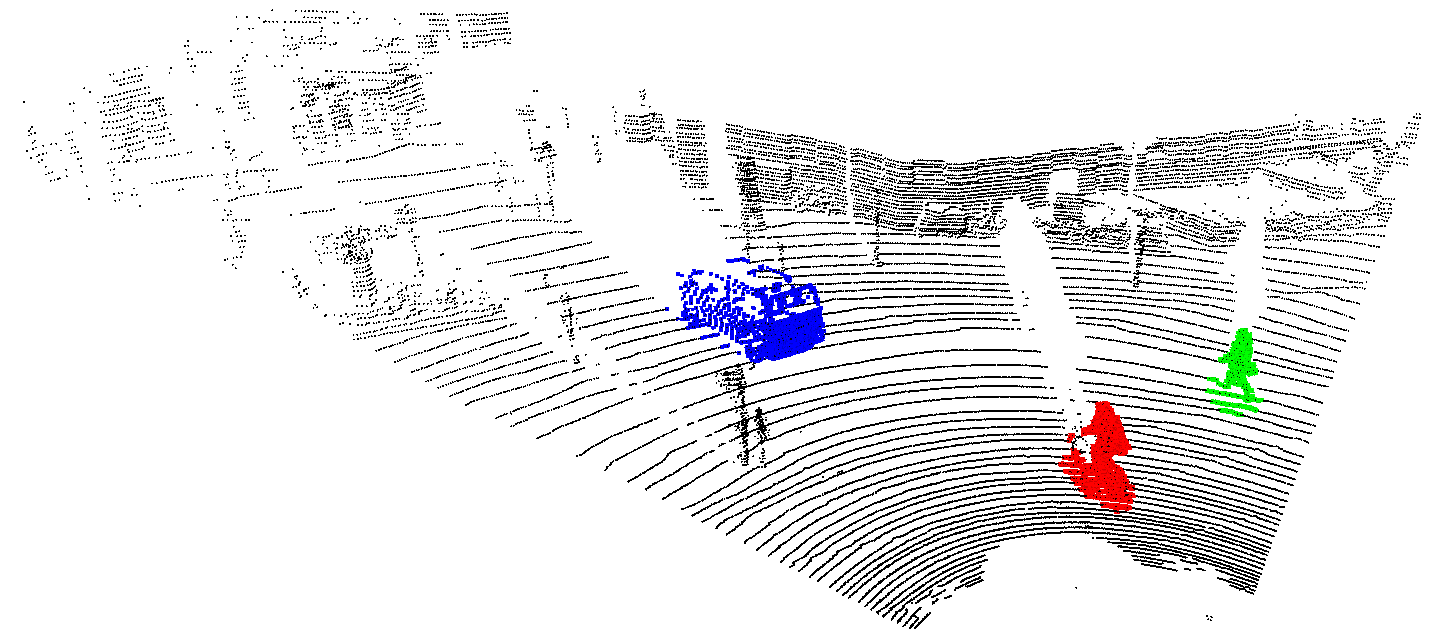}
      \caption{proposed \textit{PointSeg}}
    \end{subfigure}
    \begin{subfigure}{0.3\textwidth}
      \includegraphics[width=1\textwidth]{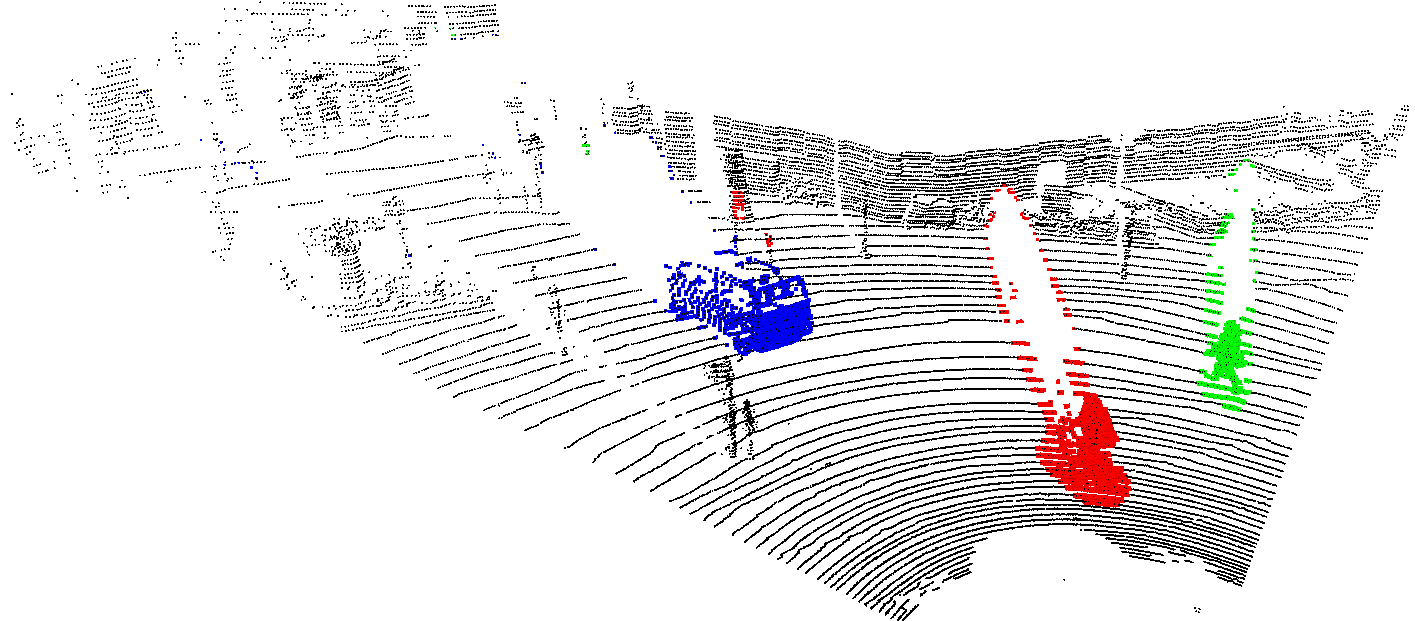}
      \caption{\textit{SqueezeSeg} \cite{squeezeseg}}
    \end{subfigure}
  \caption{
  The first column shows the input spherical data and its corresponding 3D point cloud data.
  The second column shows the predicted mask results and back projection results of \textit{PointSeg} proposed in this paper.
  The third column is the related results predicted by \textit{SqueezeSeg} \cite{squeezeseg}.
  Cars, cyclists and pedestrians are shown in blue, red and green.
  The details of the network process are shown in Fig. \ref{fig:structure}.}
  \label{fig:pipeline}
\end{figure*}

To solve these problems, we design a light network architecture for the road-object segmentation task,
which is called \textit{PointSeg}. 
The pipeline is shown in Fig. \ref{fig:pipeline}.
Our network predicts a point-wise map on the spherical image and transforms this map back to 3D space.
\subsection{Contributions}
To achieve compatible real-time performance,
we take the light-weight network \textit{SqueezeNet} \cite{squeezenet} as our root structure like \textit{SqueezeSeg} \cite{squeezeseg}.
Then we take several ideas from the state-of-the-art RGB semantic segmentation methods like PSPNet\cite{DBLP:journals/corr/ZhaoSQWJ16}
and apply them in our network together to achieve the state-of-the-art performance.
Because 3D point cloud data is naturally sparse and large, it is arduous to build real-time semantic segmentation task.
As \textit{SqueezeSeg} \cite{squeezeseg}, we solve this problem by transforming the point cloud data into a spherical image and make \textit{PointSeg} accept the transformed data.

Generally, we propose a fast semantic segmentation system for autonomous driving with the following features:
\begin{itemize}
  \item The model is quite light-weight. We extend the basic light-version network \textit{SqueezeNet} \cite{squeezenet} and \textit{SqueezeSeg} \cite{squeezeseg} with the new feature extract layers to improve the balance between accuracy and computation efficiency in 3D semantic segmentation task. 
  \item Our network can be applied directly on the autonomous driving system with an efficient interface and easy implementation with basic deep learning unit.
\end{itemize}

\section{Related works}
In this section, We discuss some recent approaches about semantic segmentation network structure, deep learning in the 3D point cloud data, bounding box detection tasks and semantic segmentation tasks.
\subsection{High Quality Semantic Segmentation for Image}
FCN \cite{long2015fully} was the pioneering method for semantic segmentation based on deep learning. It replaced the last fully-connected layers in the classification task with convolution layers.
Recent approaches like DeeplabV3 \cite{deeplabv3} used a dilated convolutional layer \cite{dilate_layer} and the conditional random field (CRF) \cite{CRF} to improve the prediction accuracy. SegNet \cite{SegNet} used an encoder-decoder architecture to fuse the feature maps from the deep layer with spatial information from lower layers. Other approaches like ICnet \cite{ICNet}, RefineNet \cite{ RefineNet} took multi-scale features into consideration and combined image features from multiple refined paths. Among those methods, Networks like Deeplabv3 and FCN were compelled to increase the performance. SegNet and ICNet are able to achieve real-time performance. Although, they have a big improvement in speed or accuracy, these methods still have some influence on the other side.

\subsection{Convolutional neural networks with 3D point cloud data}

3D data has sufficient features and attracts much research attention. With the rapidly developing of deep learning, many methods apply convolutional neural networks (CNN) on the 3D point cloud data directly. 3DFCN \cite{3DFCN} and VoxelNet \cite{VoxelNet} used a 3D-CNN \cite{DBLP:journals/corr/Graham15} to extract features from width, height and depth simultaneously.
MV3D \cite{DBLP:journals/corr/ChenMWLX16} fused multi-perception from a bird's-eye view, a front view and a camera view to obtain a more robust feature representation. In addition, some works \cite{SegMap2018} considered the representation of three-dimensional data itself and divided it into voxels to undertake features such as intensity, distance, local mean and disparity. Although all of the above methods have achieved a good accuracy, they still cost too much time in computation which limited their applications in real-time tasks. In this paper, we are aiming to improve the real-time performance and keep a good accuracy at the same time.

\subsection{Segmentation for 3D Lidar point cloud data}
Previous works proposed several different algorithms for plane extractions from 3D point clouds, such as RANSAC-based (random sample consensus) \cite{schnabel2007efficient} methods and region-grow-based methods \cite{lin2000unseeded}.
However, RANSAC requires much computation on random plane model selection. Region-grow-based methods, depending on the manually designed threshold, are not adaptive. Other traditional approaches based on clustering algorithms just realized the segmentation work but not pixel-wise region classifications.

Recently, researchers started focusing on the semantic segmentation of 3D Lidar point cloud data. PointNet \cite{DBLP:journals/corr/QiSMG16} explored a deep learning architecture to do the 3D classification and segmentation on raw 3D data. However, it only works well in indoor.
Also {Dub{\'e}} \cite{incremental2018} explored an incremental segmentation algorithm, based on region growing, to improve the 3D task performance. However, real-time performance is still challenging.
\textit{SqueezeSeg} \cite{squeezeseg} is similar with our task which used the \textit{SqueezeNet} \cite{squeezenet} as the backbone and performed compatible results. However, it only referred the CRF to improve the performance in the predicted 2D spherical masks, which could lose location information in the 3D space.
Without considering the 3D constraints in the original point cloud, the results of \textit{SqueezeSeg} is extremely limited by this CRF post-process.

\section{Method}
\label{sec:method}
We introduce the generation of spherical images and key features of the network structure in this section. Network structures and parameters are also included at the end of the section.

\subsection{Spherical image generation from point cloud}
\begin{figure}[!h]
  \centering
  \includegraphics[width=0.45\textwidth]{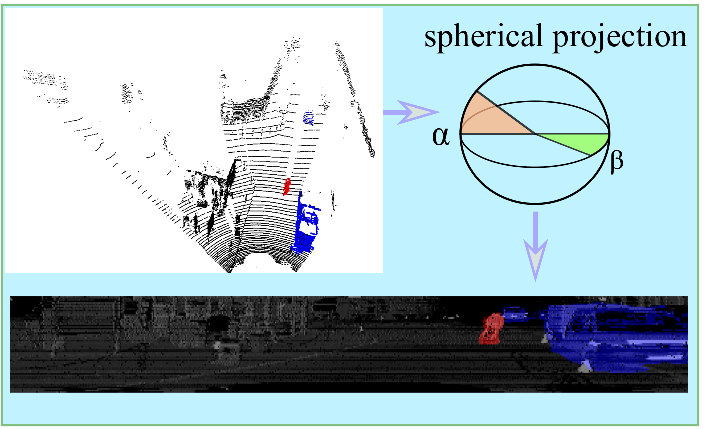}
 \caption{The spherical projection process from a point cloud to a dense spherical image as \textit{SqueezeSeg} \cite{squeezeseg}. The colored masks are cropped from ground truth boundary boxes of the KITTI dataset \cite{Kitti}.}
  \label{fig:spherical}
\end{figure}

\begin{figure*}[!h]
  \centering
    \includegraphics[width=0.9\textwidth]{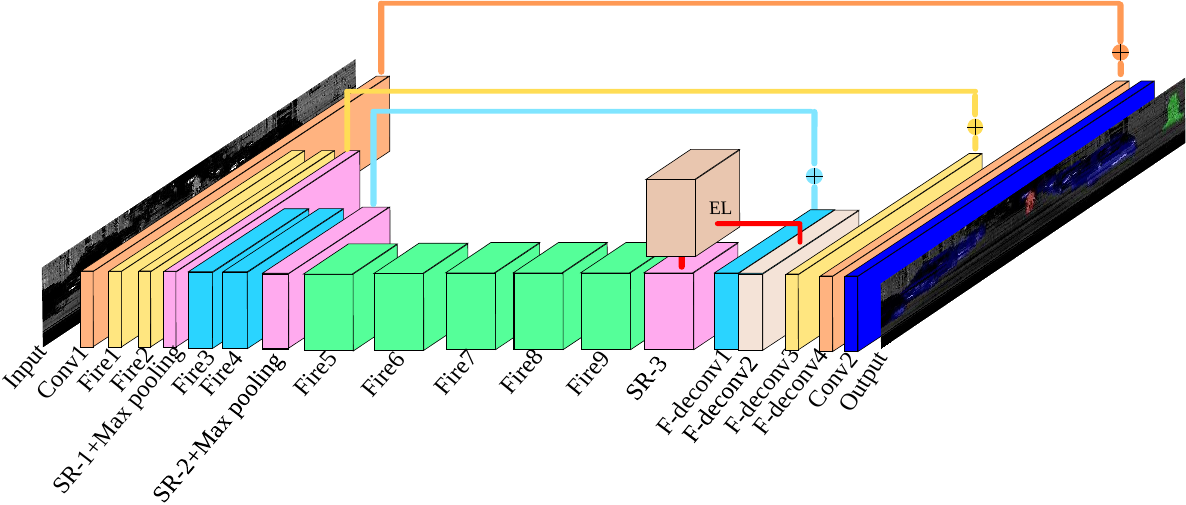}
  \caption{
  The network structure of \textit{PointSeg}.
  Our network is based on the famours light-weight strucure \textit{SqueezeNet} \cite{squeezenet} and \textit{SqueezeSeg} \cite{squeezeseg}.
  Several ideas from the state-of-the-art RGB semantic segmentation methods are considered,
  which improve both the efficiency and accuracy on this 3D task.
  \textit{Fire} is the fire layer as \textit{SqueezeNet}.
  \textit{EL} means the enlargement layer and \textit{SR} is the squeeze reweighting layer.
  }
  \label{fig:structure}
\end{figure*}

A 3D Lidar point cloud is often stored as a set of Cartesian coordinates $(x, y, z)$. We can also easily obtain the extra feature, such as RGB values (if Lidar has been calibrated with a camera) and intensities. However, the 3D Lidar point cloud is usually sparse and huge. Therefore, transforming it into voxels and then feeding voxel representations into a 3D-CNN \cite{DBLP:journals/corr/Graham15} would be computationally inefficient and memory-consuming because many voxels would be empty. To solve this problem, we transform the Lidar point cloud data by spherical projection, as the same as the \textit{SqueezeSeg} \cite{squeezeseg}, to achieve a kind of dense representation as:

\begin{equation}
\alpha = arcsin(\frac{z}{\sqrt{x^2+y^2+z^2}}),  \bar{\alpha}=\lfloor\frac{\alpha}{\Delta\alpha}\rfloor
\label{equ:azimuth},
\end{equation}
\begin{equation}
\beta =arcsin(\frac{y}{\sqrt{x^2+y^2}}),
\bar{\beta}=\lfloor\frac{\beta}{\Delta\beta}\rfloor
\label{equ:zenith},
\end{equation}
where $\alpha$ and $\beta$ are the azimuth and zenith angles, as shown in Fig. \ref{fig:spherical}; $\Delta \alpha$ and $\Delta\beta$ are the resolutions which can generate a fixed-shape spherical image; and $\bar{\alpha}$ and $\bar{\beta}$ are indexes which set the positions of points on the 2D spherical image.
After applying Equation \ref{equ:azimuth} and Equation\ref{equ:zenith} on the point cloud data, we obtain an array as $H\times W\times C$.
Here, the data is generated from Velodyne HDL-64E LiDAR with 64 vertical channels. Therefore, we set $H = 64$. Considering that in a real self-driving system most attentions are focusing on the front view, we filter the dataset only to consider the front view area $(-45^\circ, 45^\circ)$ and discretize it into 512 indexes, so $W$ is 512. $C$ is the channel of input. In our paper, we consider $x, y, z$ coordinates, intensity for each point and distance $d = \sqrt{x^2+y^2+z^2}$ as five channels in total. Therefore, we can obtain the transformed data as $64\times 512\times 5$. By this transformation, we can input it into traditional convolutional layers.

We directly extract features from the transformed data, which has dense and regular distribution. The time cost is dramatically reduced compared with taking raw 3D point cloud as inputs.

\subsection{Network structure}
\label{sec:net_struct}

The proposed \textit{PointSeg} has three main functional layers: (1) fire layer (from \textit{SqueezeNet} \cite{squeezenet}), (2) squeeze reweighting layer and (3) enlargement layer. The network structure is shown in Fig. \ref{fig:structure}.

\subsubsection{Fire layer}
\label{sec:firelayer}
\begin{figure}[!t]
 \centering
    \includegraphics[width=0.5\textwidth]{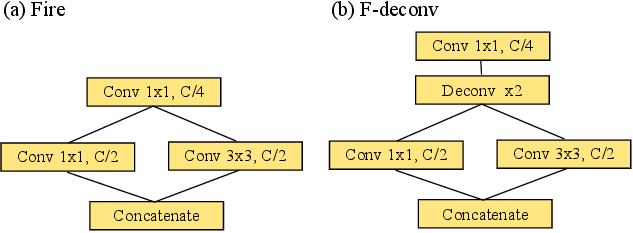}
 \caption{
 (a) is the fire model in the downsampling process as \textit{SqueezeNet} \cite{squeezenet}.
 (b) is the f-deconv model in the upsampling process as \textit{SqueezeSeg} \cite{squeezeseg}.
 }
 \label{fig:specific_net}
\end{figure}

Assessing \textit{SqueezeNet} \cite{squeezenet}, we find that its fire unit can construct a light-weight network which can achieve similar performance as AlexNet \cite{AlexNet} but costing far fewer parameters than AlexNet. Therefore, we take the fire module as our basic network unit. We follow \textit{SqueezeNet} to construct our feature extraction layer, which is shown in Fig. \ref{fig:structure} (Fire1 to Fire9). The fire module is shown in Fig. \ref{fig:specific_net} (a).
And we do not implement the \textit{MobileNet} \cite{Mobilenets}, \textit{ShuffleNet} \cite{ShuffleNet}. Because both of them can not set different $stride$ in $height$ and $width$ in the process which will influence the accuracy greatly, if we downsample the same times in \textit{height}. During the feature extraction downsampling process, we use the left one to replace the common convolutional layer. The fire module contains one squeeze module and one expansion module. The squeeze module is a single $1 \times 1$ convolution layer which compresses the model's channel dimensions from $C$ to $1/4C$. $C$ is the channel number of the input tensor. The expansion module with one $1\times 1$ convolutional layer
and one $3 \times 3$ convolutional layer help the network to achieve more feature representations from different kernel sizes.
We replace the deconvolutional layers with F-deconv like \textit{SqueezeSeg} \cite{squeezeseg} as shown in Fig. \ref{fig:specific_net} (b).

\subsubsection{Enlargement layer}
\label{sec:el_layer}
\begin{figure}[!t]
 \centering
    \includegraphics[width=0.5\textwidth]{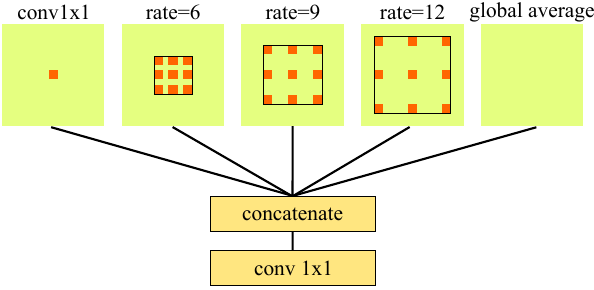}
    \caption{The structure of the enlargement layer. Here rate means the number of skipped nodes between two sampled ones. The orange points represent the sampled nodes from related feature outputs.}
 \label{fig:enlargement}
\end{figure}

Pooling layers are set to expand the receptive field and to discard the location information to aggregate the context information.
However, location information is kind of indispensable for semantic segmentation tasks.
So, here in \textit{PointSeg}, we reduce the number of pooling layers used to keep more location information. To solve this problem, instead of using the pooling layer to get a large receptive field, we deploy a dilated convolutional layer to enlarge the receptive field after Fire9 and SR-3. Similar to Atrous Spatial Pyramid Pooling (ASPP) \cite{deeplab}, we use different rates of the dilated convolutional layer to get multi-scale features at the same time. The structure of the enlargement layer is shown in Fig. \ref{fig:enlargement}.

One $1 \times 1$ convolutional layer and one global average layer are also added in the enlargement layer. Because the shape of the input feature is $64\times 64$, we set the rates as 6, 9 and 12 in three dilated convolutional layers respectively. Benefitting from this structure, we can avoid that too many zeros are added between two nodes in the traditional dilated convolutional layer and get more neighboring information. After concatenating them together, the $1\times 1$ convolutional layer is used to compress the channel from the original size to $1/4$ to avoid too much time cost in computation.

\subsubsection{Squeeze reweighting layer}
\label{sec:reweight_layer}
To obtain a more robust feature representation as efficient as possible, here we propose a reweighting layer to tackle this issue, which gets the idea from Squeeze-and-Excitation Networks \cite{DBLP:journals/corr/abs-1709-01507} and exploit channel dependencies efficiently.

We simply use a global average pooling layer to obtain the squeeze global information descriptor.
To calculate all elements $\chi$ through spatial dimensions $H\times W$ from $C$ channels, we have:

\begin{equation}
\chi_n = \frac{1}{H\times W}\sum_{i=1,j=1}^{H,W}p_n(i,j),n\in\{1,2,\cdots, C\}.
\end{equation}

\begin{figure}[!t]
  \centering
    \includegraphics[width=0.5\textwidth]{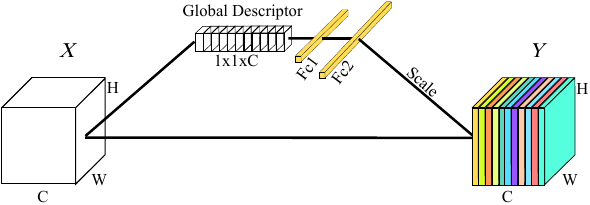}
  \caption{
  The structure of Squeeze reweight layer.
  After generated by the former feature maps,
  the \textit{Global Descriptor} will reweigh the channel-wise feature.
  }
  \label{fig:reweighting}
\end{figure}

As shown in Fig. \ref{fig:reweighting}, the channel-wise representations, with a shape of $1\times 1 \times C$, are expressive for the whole feature maps. We use two fully connected layers to generate channel-wise dependencies (called Scale in Fig. \ref{fig:reweighting}). To reweight the channel dependencies, $Y$ is formulated as

\begin{equation}
Y_n = X_n \cdot S_n, n\in\{1,2,\cdots, C\}.
\end{equation}

where $X_n=sigmoid(\chi_n)$, $S_n$ donates the output of fully connected layer as shown in Fig.    \ref{fig:reweighting}. Here we use sigmoid function to map $\chi_n$ between 0 and 1. After the squeeze reweight layer, the channel weights can be adapted to the feature representation with its global information descriptors.

\subsubsection{Details in the network}
\label{sec:detail_net}
Considering that information on height are quite limited with the input shape as $64 \times 512 \times 5$, we only do the downsampling process along the width axis and keep the same dimensions as the input features along the height axis. To get the original scale resolution for point-wise prediction, we use a deconvolutional layer to upsample the feature maps from output features of the squeeze reweighting layer (SR-3) and the enlargement layer (EL) as shown in Fig. \ref{fig:structure}).
Although we can get a large receptive field from the enlargement layer, this layer is not used any more in other layers because it will increase the parameters amounts. We also only add three squeeze reweight layers (from SR1 to SR3) before each pooling layer to help each fire block learn more robust features and reduce memory cost.
Because the feature of the enlargement layer is extracted from a different receptive field, we concatenate the outputs from enlargement layer and SR layer after F-deconv1 (as shown in Fig. \ref{fig:structure}). Skip connections are used to fuse low-level features from layers with high-level features in the other F-deconv layers. To reduce computation costs, we use $add$ instead of $concat$ in these skip operations.

\section{Experimental Evaluation}
All experiments are performed on a workstation with a single 1080ti GPU under CUDA 9 and CUDNN V7. During training, we set the learning rate as 0.001 and use the Adagrad \cite{ada} as the optimizer. We set the batch size as 32 to train the whole network for 50000 steps in around three hours. Our code is available at \url{https://github.com/ywangeq/PointSeg.git}

\begin{table*}[!t]
\caption{The results of ablation study and comparision with the state-of-the-art method}
\centering
\begin{tabular}{@{}l|lll|lll|lll@{}}
\toprule
          \multirow{2}{*}{$\%$ }   & \multicolumn{3}{c}{Car}            &  \multicolumn{3}{c}{Pedestrian}    &  \multicolumn{3}{c}{Cyclist}   \\

                               & Precision     & Recall     & IOU   & Precision     & Recall     & IOU   & Precision     & Recall     & IOU   \\
           \hline
           downsampling-4              & 63.7 & 91.3 & 62.9  & 13.0  & 19.5 & 00.8   & 20.4 & 54.7 & 18.9  \\
           downsampling-3              & 61.1 & 90.0 & 60.0  & 24.6  & 23.3 & 18.1   & 23.8  & 61.1  & 23.7  \\
           \hline
           EL-layer-(3,5,8)             & 71.6 & 94.2 & 67.5   & 16.7 & 18.7 & 14.3   & 38.1 & 55.6 & 66.3  \\
           EL-layer-(4,8,12)            & 71.1 & 90.1 & 66.3   &37.6 &  23.3 & 18.1   & 27.8  & 61.1 & 23.7 \\
           EL-layer-(6,9,12)            & 70.5 & 93.4 & 61.8  & 33.9 & 32.0  & 19.8   & 34.7 & 55.0 & 33.1  \\
           \hline
           reweight-down            & 69.2 & 92.7 & 65.7  & 34.7 & 28.8 & 18.7  &33.7 & 56.5 & 26.7 \\
           reweight-up            & 53.2 & 82.7 & 49.5  & 12.8 & 11.8 & 13.6  & 14.3 & 38.9 & 24.0  \\
           reweight-down/up  & 63.6 & 95.2 & 61.6  & 29.3 & 23.2 & 14.9  & 26.1 & 45.8 & 20.0   \\
           \hline
           \textit{SqueezeSeg}-(w/ CRF) \cite{squeezeseg}  & 66.7    & 95.4     & 64.6    & 45.2 & \textbf{29.7} & 21.8 & 35.7 & 45.8 & 25.1   \\
           \textit{SqueezeSeg}-(w/o CRF)\cite{squeezeseg}  & 62.7 & \textbf{95.5} & 60.9  & \textbf{52.9} & 28.6 & \textbf{22.8}  & 35.2 & 51.1 & 26.4 \\
           \textit{pointseg}                              & \textbf{74.8} & 92.3 & \textbf{67.4} & 41.4 & 29.3 & 19.2  & \textbf{41.4} & \textbf{59.7} & \textbf{32.7}  \\
           \hline
           \textit{pointseg}-(w/ RANSAC)                  & 77.2 & 96.2 & 67.3 & 48.6 & 29.4 & 23.9 & 46.3 & 63.3 & 38.7 \\
           \bottomrule
\end{tabular}
\label{table:all_result}
\end{table*}

\subsection{Dataset and evaluation metrics}
We train our model on a published dataset\footnote{\url{www.dropbox.com/s/pnzgcitvppmwfuf/lidar_2d.tgz}} from \textit{SqueezeSeg} \cite{squeezeseg} which converts Velodyne data from the KITTI 3D object detection dataset \cite{Kitti}. It was split into a training set with around 8000 frames and a validation set with around 2800 frames.
The evaluation precision, recall and IoU are defined as follows:
$$
P_n = \frac{|\rho_n \bigcap G_n |}{|\rho_n|},
R_n = \frac{|\rho_n \bigcap G_n |}{|G_n|},
IoU_n = \frac{|\rho_n \bigcap G_n|}{\rho_n \bigcup G_n}
$$
where $\rho_n$ is the predicted sets that belong to class-n, and $G_n$ is the ground-truth sets.

\subsection{Ablation study}
Our network is mainly built based on the excellent \textit{SqueezeNet} \cite{squeezenet}.
Compared with another very similar work \textit{SqueezeSeg} \cite{squeezeseg}, we improve the network structure from several parts and here we show the ablation study results respectively. All the results are shown in Table \ref{table:all_result} as percentage.

\subsubsection{Downsampling times}
\label{sec:downsampling}
The original \textit{SqueezeNet} \cite{squeezenet} contains four downsampling layers.
The shape of the generated feature is ($1/16$H$\times1/16$W$\times$C) after downsampling four times.
Through this structure, the output feature will be $64\times 32\times$C without downsampling in height if our input is $64\times512\times$C.
To restore more location information as we mentioned in Section \ref{sec:el_layer},
we remove the last downsampling layer of the basic \textit{SqueezeNet}.

%

In the first part of Table \ref{table:all_result},
We compare the results of different downsampling times (downsampling-4 for four downsampling times and downsampling-3 for three downsampling times).
Note that downsampling-4 is actually as the same as the \textit{SqueezeSeg} \cite{squeezeseg} without CRF.
Here the first row shows the result of \textit{SqueezeSeg} without CRF trained by ourselves to evaluate the efficiency of our proposed strategy.
The results show that reducing the downsampling times from four to three dramatically improves the accuracy of pedestrian and cyclist, which are relatively small and easily affected by downsampling. At the same time, the predictions for cars are still comparable.
Thus we choose to downsample three times in our \textit{PointSeg}. All the ablation experiments following are also based on this downsampling-3 structure.


\subsubsection{Ablation study for enlargment layer}
\label{sec:abl_el}
Because the size of output feature (Fire9) size is 64$\times$64 in height and weight. To make enlargement layer described in Section. \ref{sec:el_layer} achieve better performance, we evaluate different rates of the enlargement layer according to previous experiences in Deeplabv3 \cite{deeplab} and hybrid dilated convolution \cite{HDC}.
%

In the second part of Table \ref{table:all_result},
we evaluate different enlargement parameters based on the downsampling-3 structure mentioned in Section. \ref{sec:downsampling}.
We only implement the enlargement layer after \textit{fire 9} as shown in Fig. \ref{fig:structure},
which increases the memory cost of the whole structure from 1.6G to 1.8G.

We also tried adding another enlargement layer after \textit{fire 4}. However, we obtain little performance improvement but a terrible increase in the memory cost.
The different rate sets have the same memory cost where the difference is that they will obtain different eyesight field.
Based on the results shown in Table \ref{table:all_result}, we choose (6,9,12) as the rate of the enlargment layer in our proposed \textit{PointSeg}.
At this stage, we notice that although the performance of the car has been improved, the performance of the pedestrian and cyclist still do not achieve the aim which we expected. A possible explanation is that the distortion and uncommon deformation from the input spherical image make the network difficult to predict those relatively small objects with similar patterns.

\subsubsection{Ablation study for reweight layer}
\label{sec:abl_reweight}
Based on the discussion in Section \ref{sec:abl_reweight},
we consider to utilize reweight layers to enhance the feature representation in channel-wise for small objects in our scenarios.
We experiment with three methods to combine the network with reweight layer. which are
(i)reweight-down: add reweighting layers at the end of each size-invariant block in the feature downsampling process like SR-1, SR-2 and SR-3 shown in Fig. \ref{fig:structure}.
(ii)reweight-up: add reweighting layers after each size-variant upsampling process, which are located after F-deconv1, F-deconv3 and F-deconv4 as shown in Fig. \ref{fig:structure}.
(ii)reweight-down/up: add reweighting layers based on the three skip connections where both the downsampling and upsampling features with the same size are combined as the input of reweighting layers.
Experiments in this section are based on the downsampling-3 structure (Section \ref{sec:downsampling}) without considering the enlargement layer (Section \ref{sec:abl_el}).
\begin{table}[]
\centering

\caption{The comparsion of runtime performance}
\begin{tabular}{@{}lll@{}}
\toprule
Methods    &           & Time(ms) \\ \midrule
\textit{SqueezeSeg} \cite{squeezeseg} & w/ CRF    & 13.5     \\\midrule
           & w/o CRF   & 8.5      \\\midrule
\textit{PointSeg}   & w RANSAC & 14       \\\midrule
           & w/o RANSAC  & 12       \\ \midrule
\textit{PointSeg} in TX2   & w/o RANSAC & 98       \\\bottomrule

\end{tabular}
\label{table:runtime}

\end{table}

\begin{figure*}[!t]
  \centering
  \includegraphics[width=1\textwidth]{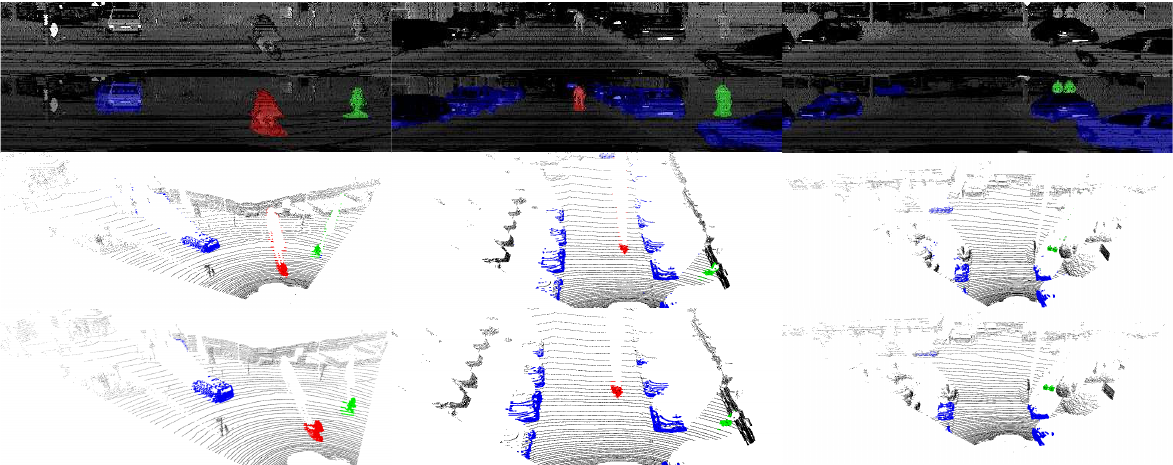}
  \caption{
  Visualizations of raw inputs, \textit{PointSeg} predictions and results after back projection with or without RANSAC refinements from up to down.
  The third row shows results which are projected back without RANSAC and the forth row shows resutls wich are projected back without RANSAC.}
  \label{fig:ras}
\end{figure*}

In Table \ref{table:all_result}, both reweight-down and reweight-down/up show better performance than the baseline downsampling-3 structure.
According to experiments about reweight layers, we find that most of the key features for this task are coming from the downsampling process which is the best time to reweight the layer weight.
If we implement the reweight layer where the global descriptor is generated from the upsampled feature as reweight-up, the results decrease obviously because reweighting feature weight from deconvolution layer may add noise on feature locations.
Basically, We add the reweight layers only at downsampling process as reweight-down.
We also tried to add the reweight layer after each layer in the downsampling process. However, the accuracy improves slightly but costing a lot of extra time.
\subsubsection{Comparison with \textit{SqueezeSeg}}
Finally, We compare our results with \textit{SqueezeSeg} \cite{squeezeseg}, which is summarized in the third part of Table \ref{table:all_result}.
Our results for the pedestrian are comparable with \textit{SqueezeSeg} (without CRF) and show great improvement for car and cyclist.
Our system cost 12 ms per frame in our workstation during the forward process with 2G memory cost.
The comparison of runtime performance is shown in Table \ref{table:runtime}.
During the back projection process from the mask on the spherical image to the point cloud data,
we use random sample consensus (RANSAC) to do the outlier remove.
The operation can help our proposed \textit{PointSeg} obtain a refined segmentation result as shown in Fig. \ref{fig:ras}, and only cost around 2ms extra time.
The evaluation result of \textit{PointSeg} aided with RANSAC is shown in the last row of Table \ref{table:all_result} which we do not compare with \textit{SqueezeSeg} due to the randomness of RANSAC.

%
%

%

\section{Conclusion}
In this paper,
we improved the feature-wise and channel-wise attention of the network to get the robust feature representation,
which shows an essential improvement in 3D semantics segmentation task from spherical images.
The proposed \textit{PointSeg} system can be directly applied in autonomous driving systems and implemented in embedded AI computing device with limited memory cost.

Besides, we focus on the computation ability and memory cost during the implementation.
Therefore, our approach can achieve a high accuracy at real-time speeds, and spare enough space and computation ability for other tasks in driving systems.
After an efficient RANSAC post-process, our method dramatically oversteps the state-of-the-art method.


In the proposed PointSeg, the performance of the small object, like the pedestrian, still do not achieve a very high level. A possible explanation is that much useful information had lost during the downsampling process even we only took three times because of the quite small shape of those original objects in the spherical image. We leave those as the future work.


\addtolength{\textheight}{-1cm}   
\clearpage
\bibliographystyle{IEEEtran}
\bibliography{wang19icra}  








\end{document}